\documentclass[lettersize,journal]{IEEEtran}
\usepackage{multirow}
\usepackage{multicol}
\usepackage{makecell}
\usepackage{amsmath}
\usepackage{amssymb}
\usepackage{cite}
\usepackage{tikz,graphics,color}
\usepackage{colortbl}
\usetikzlibrary{shadows,patterns,shapes,arrows,decorations.pathmorphing,backgrounds,positioning,fit,plotmarks,calc,spy,petri,topaths,matrix}
\usetikzlibrary{arrows.meta}
\usepackage{pgfplots}
\pgfplotsset{compat=1.18}
\usepackage{pgfplotstable}
\usepgfplotslibrary[groupplots]
\pgfplotsset{compat=1.18,
    /pgfplots/ybar legend/.style={
    /pgfplots/legend image code/.code={%
       \draw[##1,/tikz/.cd,yshift=-0.25em]
        (0cm,0cm) rectangle (4pt,0.6em);},
   },
   legend style={
    /tikz/execute at begin node = \scriptsize\itshape
  },
}
\usepackage{pifont}
\usepackage{wrapfig}
\usepackage{enumitem}
\usepackage{url}
\usepackage{balance}

\definecolor{PastelRose}      {RGB}{251,180,174} 
\definecolor{PastelSkyBlue}   {RGB}{179,205,227} 
\definecolor{PastelMintGreen} {RGB}{204,235,197} 
\definecolor{PastelLavender}  {RGB}{222,203,228} 
\definecolor{PastelPeach}     {RGB}{254,217,166} 
\definecolor{PastelButter}    {RGB}{255,255,204} 
\definecolor{PastelBeige}     {RGB}{229,216,189} 
\definecolor{PastelPink}      {RGB}{253,218,236} 
\definecolor{PastelLightGrey} {RGB}{242,242,242} 

\definecolor{darkgoldenrod}{rgb}{0.72, 0.53, 0.04}
\definecolor{blizzardblue}{rgb}{0.67, 0.9, 0.93} 
\definecolor{cadmiumgreen}{rgb}{0.0, 0.42, 0.24} 
\definecolor{satPointsColor}{HTML}{D7191C}
\definecolor{unsatPointsColor}{HTML}{FDAE61}
\definecolor{timedOutPointsColor}{HTML}{ABDDA4}
\definecolor{rendering}{HTML}{2B83BA}
\definecolor{deeperorange}{RGB}{205,85,0}
\definecolor{deeperBlue}{RGB}{100, 149, 237}  
\definecolor{deeperRed}{RGB}{220, 20, 60}     
\definecolor{deeperYellow}{RGB}{238, 220, 130} 
\definecolor{deeperGreen}{RGB}{34, 139, 34}   
\definecolor{deeperPurple}{RGB}{104, 34, 139} 
\definecolor{deeperGray}{RGB}{105, 105, 105}  
\definecolor{richYellow}{RGB}{204, 204, 0}  

\usepackage{comment}
\usepackage{verbatim} 
\usepackage{graphicx}
\usepackage{pifont}
\usepackage[export]{adjustbox}
\usepackage{pgf-pie}
\usepackage{tikz}
\usepackage{tkz-graph}
\usepackage{hyperref}
\usepackage{threeparttable} 
\hypersetup{
    colorlinks,
    linkcolor={red!40!black},
    citecolor={blue!40!black},
    urlcolor={black}
}
\usepackage[colorinlistoftodos]{todonotes}

\begin{document}

\title{SAEM: \underline{S}tage-\underline{A}ware \underline{E}xpert \underline{M}anagement for Memory-Efficient MoE Inference in Chain-of-Thought Reasoning}

\author{Yujie Zhang, Bin Gao, and Tulika Mitra
\thanks{This work was supported by the National Research Foundation, Singapore, under its Competitive Research Program Award NRF-CRP23-2019-0003 and the Ministry of Education, Singapore, under Tier 3 grant MOE-MOET32024-0003. 
Yujie Zhang, Bin Gao, and Tulika Mitra are with the School of Computing, National University of Singapore, Singapore (e-mail: zyujie@comp.nus.edu.sg; bingao@nus.edu.sg; tulika@comp.nus.edu.sg).

A preliminary version of this work was \textbf{accepted for publication at the 63rd ACM/IEEE Design Automation Conference (DAC 2026)}, Long Beach, CA, USA; this article extends it with a detailed treatment of reasoning stage boundary detection, evaluation on AIME 2024 and GPQA-Diamond, and a prediction-oracle upper-bound analysis.
}
}




\maketitle
\thispagestyle{empty}

\begin{abstract}
Chain-of-thought (CoT) prompting improves LLM reasoning by decomposing complex problems into intermediate steps, but its sequential nature increases decoding latency and memory usage. Mixture-of-Experts (MoE) models scale capacity through sparse expert activation, yet their full expert weights often exceed GPU memory and require costly GPU--CPU transfers. 
Existing runtimes treat all tokens uniformly, overlooking a key structural property of CoT traces: consecutive reasoning stages exhibit coherent and predictable expert activation patterns. Ignoring this stage-level regularity leads to inefficient caching and unnecessary data movement.
We propose SAEM, a stage-aware MoE inference runtime that detects reasoning stage boundaries and exploits stage-level activation coherence to guide expert placement.
SAEM combines stage-aware caching, expert-aligned token repacking, and in-situ CPU execution to reduce data transfer and kernel fragmentation.
On mathematical and scientific reasoning workloads, SAEM achieves an average 1.33$\times$ throughput improvement over the strongest state-of-the-art caching and offloading baselines under constrained GPU memory, rising to 1.54$\times$ when calibration data matches the workload, demonstrating the effectiveness of stage-aware, locality-driven MoE inference for CoT reasoning.
\end{abstract}


\section{Introduction}
\label{sec:saem_intro}
\IEEEPARstart{L}{arge} language models (LLMs) have demonstrated strong reasoning capabilities through chain-of-thought (CoT) prompting, which decomposes complex problems into structured intermediate steps~\cite{wei2022chain,lyu2023faithful,yeo2025demystifying}. This approach is particularly effective in mathematically demanding domains, where explicit step-by-step reasoning substantially improves accuracy. However, CoT decoding incurs considerable computational cost. The autoregressive generation of long reasoning traces increases both inference latency and memory pressure; on challenging reasoning benchmarks, such traces frequently span hundreds or thousands of tokens, and evaluations of long-output reasoning models often permit generation budgets of up to 32,768 tokens~\cite{guo2025deepseek}. Consequently, reasoning trace length tends to increase with task difficulty, further amplifying decoding cost and memory demand.

Mixture-of-Experts (MoE) architectures offer a promising direction for scalable inference by activating only a sparse subset of experts per token, enabling large model capacity without proportionally increasing computation~\cite{guo2025deepseek,jaech2024openai}. 
Yet deploying MoE models for CoT reasoning under resource constraints remains challenging. The full set of expert weights typically exceeds available GPU memory, forcing dynamic movement of expert parameters between CPU and GPU. 
Moreover, even when GPU memory is sufficient to hold all experts, the inherent sparsity of MoE routing means that most GPU-resident experts are seldom activated, leading to significant underutilization of scarce GPU memory and undermining the benefits of full on-device placement.

Quantization and sparsity approaches~\cite{chitty2025mopeq,lu2024not} reduce computational and memory costs but often yield inconsistent performance across task domains.
Recent MoE inference runtimes instead mitigate memory limitations through token-level expert caching~\cite{eliseev2023fast,zhong2025hybrimoe,yu2025prescope,tang2024hobbit} and expert prefetching~\cite{du2024sida,xue2024moe,fang2025accurate,hwang2024pre}. Caching systems track expert usage and update the GPU cache based on recency or frequency. Prefetching complements this by predicting future expert requirements and proactively transferring expert weights to GPU memory.
These approaches, however, treat generated tokens uniformly and overlook a key structural property of CoT reasoning: consecutive reasoning stages exhibit strong semantic coherence and consistently activate predictable expert subsets. When models explore alternative solution paths, perform self-correction, or validate intermediate results, they transition between distinct reasoning modes, each associated with stable and stage-specific expert activation patterns. Token-level expert management cannot capture these coarse-grained regularities, resulting in unnecessary GPU--CPU data movement.

We introduce SAEM, a data-aware MoE inference runtime for resource-constrained CoT reasoning that leverages semantic structure rather than token-level uniformity.
SAEM builds on a key empirical observation: reasoning stages marked by discourse transitions (e.g., ``alternatively,'' ``instead,'') exhibit coherent expert activation patterns that remain stable within stages but shift predictably across boundaries. This stage-level coherence provides actionable guidance for cache management that token-level approaches cannot exploit.

SAEM employs three coordinated mechanisms to optimize inference efficiency. First, it uses lightweight pattern matching to detect stage transitions in real time and aggregates expert usage statistics at stage granularity to guide cache management. This coarse-grained strategy updates GPU expert residency only when reasoning semantics change, substantially reducing data movement compared to fine-grained token-level approaches while maintaining a high cache hit rate. Second, SAEM reorganizes tokens by expert assignment through expert-aligned token repacking, converting scattered memory accesses into contiguous batches that improve GPU utilization and reduce kernel launch overhead, particularly under highly skewed expert routing. Third, SAEM executes infrequently activated experts directly on the CPU via in-situ execution, avoiding redundant PCIe transfers and preventing cache pollution from experts unlikely to be reused within the current reasoning stage. These three mechanisms address complementary aspects of the inference challenge and thus form an integrated system rather than a set of independent optimizations.

Our main contributions are as follows:
\begin{itemize}[leftmargin=*,noitemsep,topsep=0pt]
    \item We identify key empirical characteristics of expert activation during CoT reasoning in MoE inference on complex reasoning tasks, and show how these insights can guide runtime hardware resource allocation.
    
    \item We present SAEM, a stage-aware inference runtime that reduces memory overhead through coordinated expert caching, token reorganization, and selective CPU execution while maintaining high throughput under constrained GPU memory. 
    
    \item Through extensive evaluation across two MoE models, three reasoning benchmarks, and varying batch sizes and GPU cache budgets, we demonstrate that SAEM consistently outperforms state-of-the-art fine-grained expert caching and offloading baselines, with an average 1.33$\times$ throughput improvement overall and 1.54$\times$ under calibration-matched conditions.
\end{itemize}

\section{Background}
\label{sec:saem_background}
\subsection{Mixture-of-Experts Inference}
\subsubsection{Mixture-of-Experts Inference}
MoE models replace dense feed-forward layers with sparsely activated expert networks. A routing mechanism selects only a small subset of experts (e.g., top-$k$) per token, enabling substantial parameter scaling without proportional growth in computation. 
This sparse activation mechanism allows parameter growth far beyond what dense models can practically support, while keeping per-token computation relatively stable. 
At inference time, however, expert activations are often highly skewed and strongly dependent on the input domain, despite load-balancing regularization during training.
This skew amplifies the need for effective expert placement and scheduling.

\subsubsection{Resource-Efficient MoE Inference}
To address these challenges, prior systems employ various dynamic expert management strategies, including GPU-side caching~\cite{eliseev2023fast,zhong2025hybrimoe,yu2025prescope,tang2024hobbit}, CPU-side execution of non-resident experts~\cite{kamahori2024fiddler}, and sequence-level expert prediction and prefetching~\cite{zhang2025daop,xue2024moe,du2024sida}. These techniques reduce transfer overhead and improve memory utilization, but they generally operate at token-level, calibration-based, or sequence-level granularity and thus overlook the semantic structure of generated outputs. In multi-step reasoning tasks, CoT traces exhibit distinct reasoning stages with stable, predictable expert activation patterns—regularities that finer-grained management fails to exploit.

\begin{figure*}[t]
    \centering
    \includegraphics[width=0.85\linewidth]{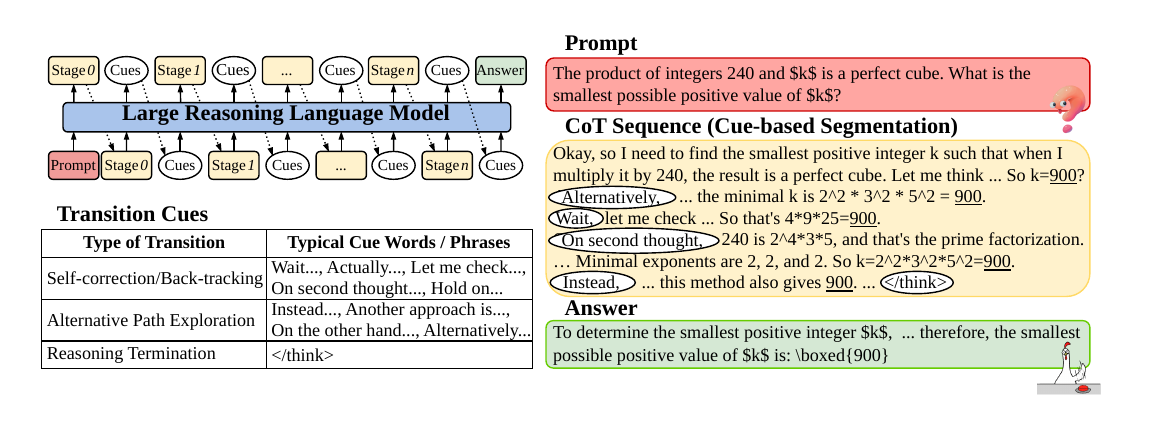}
    \caption{CoT reasoning employs transition cues to structure multi-step problem solving, facilitating self-correction and the exploration of alternative solution paths.}
    \label{fig:saem_cot_intro}
\end{figure*}


\begin{figure}[t]
    \centering
    \includegraphics[width=0.75\linewidth]{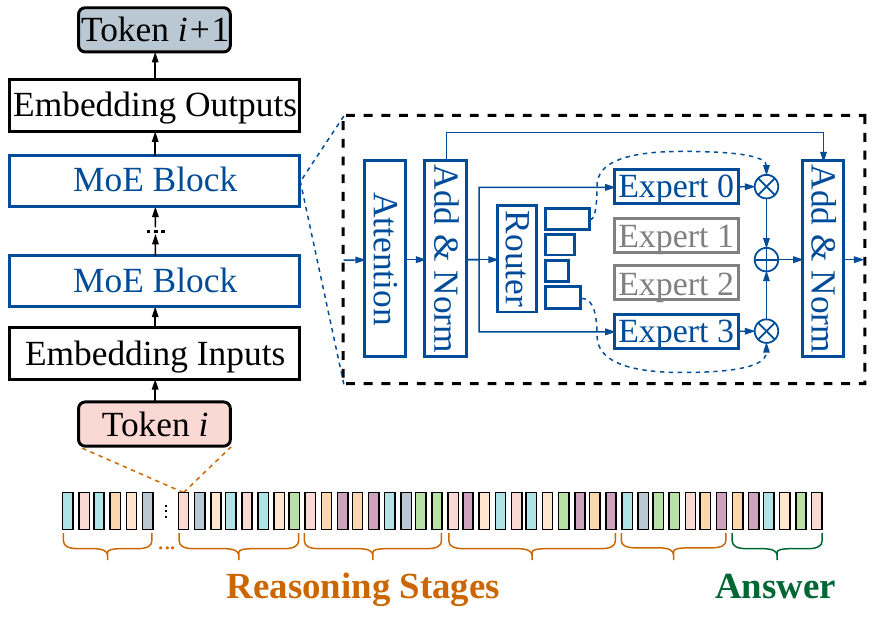}
    \caption{Illustration of CoT reasoning in a MoE model, where each token activates its top-2 selected experts.}
    \label{fig:cot_reasoning}
\end{figure}

\begin{figure}[t]
    \centering
    \includegraphics[width=0.8\linewidth]{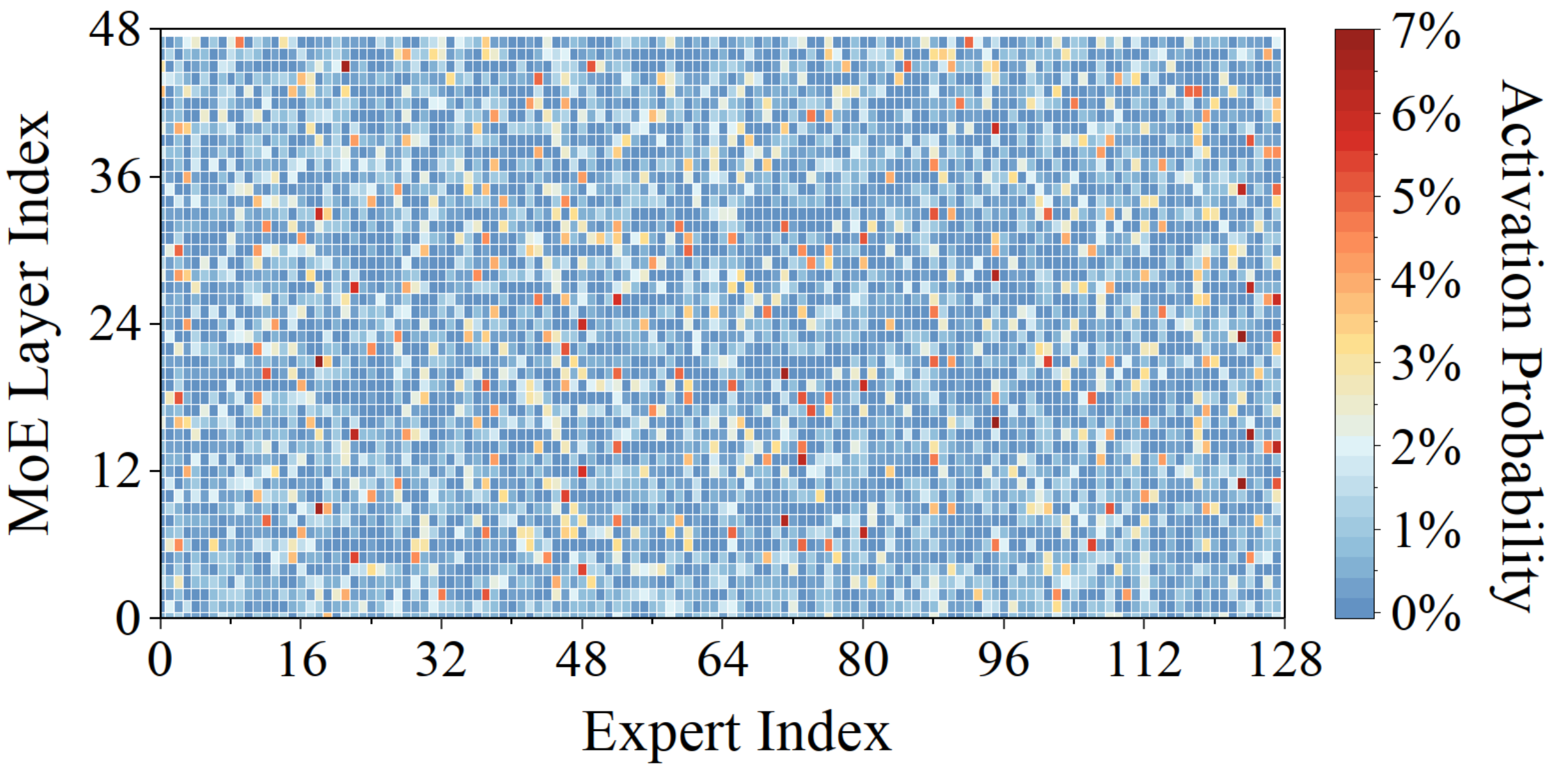}
    \caption{Layer-wise expert activation pattern in Qwen3 on MATH‑500 under CoT reasoning, averaged over 100 samples. 
    }
    \label{fig:saem_eaps_qwen3}
\end{figure}

\subsection{Chain-of-Thought Reasoning}
\subsubsection{CoT as Structured Multi-Step Reasoning}
Chain-of-thought (CoT) reasoning enables step-by-step ``thinking'' by decomposing complex problems into intermediate subgoals and deductions~\cite{jaech2024openai}. Rather than producing answers directly, models articulate their reasoning process, often involving self-correction, backtracking, and exploration of alternatives. This structured approach is widely adopted in advanced LLMs such as OpenAI o1~\cite{jaech2024openai} and DeepSeek-R1~\cite{guo2025deepseek} to improve accuracy and robustness on mathematical and logical reasoning tasks.
 
\subsubsection{Internal Structure of CoT Traces}
CoT traces are not linear streams of tokens but structured sequences composed of distinct reasoning stages. These stages are typically signaled by discourse-level transition cues such as ``alternatively,'' ``on second thought,'' or ``therefore,'' which indicate shifts in reasoning strategy, verification, or conclusion (Fig.~\ref{fig:saem_cot_intro}). These cues segment the reasoning process into semantically coherent units. 
Crucially, different stages—such as exploring alternatives, validating assumptions, or finalizing conclusions—exhibit distinct computational patterns and activate different expert subsets within the model (Fig.~\ref{fig:cot_reasoning}). This reveals exploitable structural and temporal regularities in expert usage that current MoE inference systems do not leverage.


\section{Motivation}
\label{sec:motivation}
\noindent\textbf{The Opportunity:} Conventional inference systems treat CoT decoding as a uniform token stream and manage expert placement at token granularity, overlooking semantic regularities across reasoning stages.
As shown in Fig.~\ref{fig:cot_reasoning}, different stages (exploring alternatives, validating assumptions, or finalizing conclusions) activate distinct expert subsets. Recognizing these stage transitions enables more efficient expert utilization in MoE architectures. This section presents empirical observations that motivate our design for resource-constrained MoE inference in CoT reasoning.

\noindent\textbf{{Motivation 1: Sparse and Skewed Expert Activation in CoT Reasoning Enables Targeted Caching.}}
Although MoE models are trained with load-balancing objectives to encourage uniform expert utilization~\cite{fedus2022review}, domain-specific reasoning tasks exhibit highly skewed routing behavior, resulting in sparse and imbalanced expert usage.
Fig.~\ref{fig:saem_eaps_qwen3} shows layer-wise expert activation patterns of Qwen3-30B-A3B (Qwen3) on MATH-500, averaged over 100 samples. 
Under a top-8 gating policy and assuming uniform routing across 128 experts, each expert would be expected to be selected with probability $\sim$6.25\%. 
In practice, however, only a small subset of experts consistently exceeds this baseline, while the vast majority are rarely or never selected.
This pronounced skew reveals a strong expert-selection bias: during CoT reasoning, the model repeatedly routes tokens to a narrow set of experts aligned with semantic or task-specific subspaces. Such sparsity and concentration create a clear opportunity for runtime optimization. By caching frequently activated experts and deprioritizing inactive ones, the inference system can reduce memory traffic, avoid unnecessary expert loading, and thereby improve overall computational efficiency.

\setlength\tabcolsep{4.2pt} 
\begin{table}[t]
\centering
\caption{Temporal coherence scores ($TC_{seq}$) of expert activation patterns across reasoning stages, segmented by transition cues, measured on Qwen3 and ERNIE-4.5.}
\label{tab:seq_sim_dist}
    \begin{tabular}{|c|c|c|c|}
        \hline
         & MATH-500 & AIME 2024 & GPQA-Diamond \\
        \hline
        \hline
         Qwen3 & 89.78\% & 89.50\% & 92.01\% \\
         \hline
         ERNIE-4.5 & 89.35\% & 89.87\% & 85.27\% \\
        \hline
    \end{tabular}
\end{table}

\noindent\textbf{{Motivation 2: Temporal Coherence Across Reasoning Stages Enables Predictive Scheduling.}}
Expert routing remains remarkably stable across adjacent CoT stages, with transitions indicated by linguistic or semantic cues that signal shifts in reasoning strategy, such as exploring alternatives (e.g., ``Alternatively,'', ``Instead,''). 

To formalize this observation, let $R^{p} \in \mathbb{R}^{L \times E}$ denote the expert activation matrix for stage $p$, where $L$ is the number of MoE layers and $E$ the number of experts per layer.
Each entry $R^{p}_{i,j}$ represents the activation probability of expert $j$ in layer $i$, computed as the ratio of tokens routed to that expert relative to all tokens processed by layer $i$.
To quantify stage-to-stage consistency, we compute the average layer-wise cosine similarity between the activation matrices of consecutive stages:
$sim\bigl(R^{p},R^{p+1}\bigr) = \frac{1}{L}\sum_{l=1}^{L}\cos\bigl(R^{p}_{l},\,R^{p+1}_{l}\bigr).$

Aggregating over a multi-stage reasoning trace of length $P$ yields the sequence-level temporal coherence metric:
\begin{equation}
    TC_{seq} = \frac{1}{P-1} \sum_{p=1}^{P-1} sim \left(R^{p}, R^{p+1}\right),
\end{equation}
where higher values indicate more stable routing across stage transitions.
Table~\ref{tab:seq_sim_dist} reports $TC_{seq}$ scores for Qwen3 and ERNIE-4.5 across three datasets, with an average of 89.30\%.

This coherence holds despite substantial variation in stage structure. Across Qwen3-generated reasoning traces on MATH-500, the number of stages ranges from 1 to 49 (median 5, average 8), while stage length ranges from 55 to 7129 tokens (median 271, average 484). This suggests that SAEM benefits from local expert-activation regularity between adjacent stages rather than from uniformly short or homogeneous traces. Accordingly, stage-aware scheduling can use the activation pattern of stage $p$ to proactively place experts for stage $p+1$, reducing cache updates and expert transfer overhead. Although scalability to substantially longer contexts remains future work, the same principle should apply as long as adjacent-stage coherence persists.

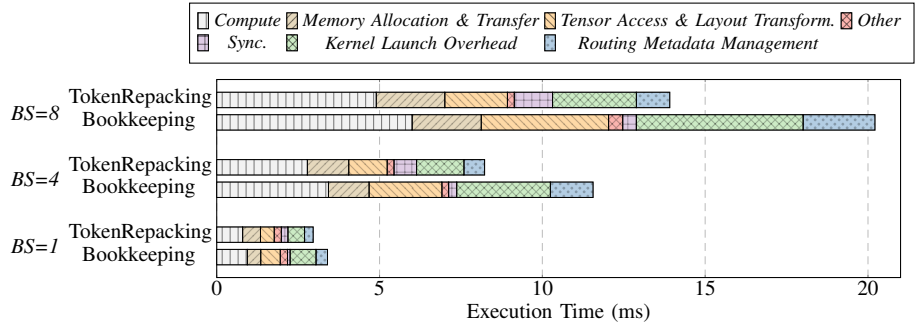
\begin{figure*}[tb]
    \centering
    \begin{tikzpicture}   
    \pgfplotstableread[col sep=tab]{data/breakdown_comp.dat}\datatable

    \begin{groupplot}[
        group style={
            group name=my plots,
            group size=1 by 1,
            xlabels at=edge bottom,
            xticklabels at=edge bottom,
        },
        ytick=data,
        yticklabels from table={\datatable}{Label},
        xmajorgrids,
        ytick pos=left,
        major tick length=1pt,
        grid style=densely dashed,
        enlarge y limits=0.13,
        y tick label style={yshift=-1.68pt,font=\footnotesize,rotate=0,text height=2pt,text width=1.85cm,align=center,inner sep=2pt,outer sep=1pt},
        xlabel={Execution Time (ms)},
        x tick label style={font=\footnotesize,rotate=0,text height=2pt,inner sep=2.5pt,outer sep=2pt},
        ylabel style={font=\footnotesize,rotate=270,align=center,text height=0.1pt,inner sep=0pt,outer sep=0pt},
        xlabel style={font=\footnotesize, align=center, text height=1pt,inner sep=1pt,outer sep=1.5pt},]

        \nextgroupplot[width=1.2\columnwidth,, height=4.2cm,
                        xbar stacked,
                        bar width=5.8pt,
                        xmin=0,
                        xmax=21,
                        xtick={0,5,10,15,20,25},
                        xticklabels={0,5,10,15,20,25},
                        typeset ticklabels with strut,
                        legend columns=4,
                        legend style={at={(0.49,1.08)},anchor=south,font=\footnotesize,
                        row sep=-2pt,
                        },
                        legend image code/.code={%
                           \draw[yshift=-0.25em]
                            (0cm,0cm) rectangle (4pt, 0.6em);}]
                            
                    \addplot[fill=PastelLightGrey,draw=black,line width=0.2pt, postaction={pattern=vertical lines, opacity=0.5}] table[y=Clusters,x=Compute] {\datatable};
                    \addlegendentry{Compute};
                    
                    \addplot[fill=PastelBeige,draw=black,line width=0.2pt, postaction={pattern=north east lines, opacity=0.5}] table[y=Clusters,x=Memory] {\datatable};
                    \addlegendentry{Memory Allocation \& Transfer};
                    
                    \addplot[fill=PastelPeach,draw=black,line width=0.2pt,postaction={pattern=north west lines, opacity=0.5}] table[y=Clusters,x=Indexing/Gather/Scatter] {\datatable};
                    \addlegendentry{Tensor Access \& Layout Transform.};  

                    \addplot[fill=PastelRose,draw=black,line width=0.2pt,postaction={pattern=crosshatch, opacity=0.5}] table[y=Clusters,x=Other] {\datatable};
                    \addlegendentry{Other};
                    
                    \addplot[fill=PastelLavender,draw=black,line width=0.2pt,postaction={pattern=grid, opacity=0.5}] table[y=Clusters,x=Sync] {\datatable};
                    \addlegendentry{Sync.};
                    
                    \addplot[fill=PastelMintGreen,draw=black,line width=0.2pt,postaction={pattern=crosshatch, opacity=0.5}] table[y=Clusters,x=Launch] {\datatable};
                    \addlegendentry{Kernel Launch Overhead};
                    
                    \addplot[fill=PastelSkyBlue,draw=black,line width=0.2pt,postaction={pattern=crosshatch dots, opacity=0.5}] table[y=Clusters,x=Bookkeeping] {\datatable};
                    \addlegendentry{Routing Metadata Management};

    \end{groupplot} 
    \draw (-2.4,0.42)  node{\footnotesize{\em BS=1}};
    \draw (-2.4,1.3)  node{\footnotesize{\em BS=4}};
    \draw (-2.4,2.18)  node{\footnotesize{\em BS=8}};
 
\end{tikzpicture}
    \caption{Operation-level breakdown of \textit{Bookkeeping} and \textit{Token Repacking} strategies in one MoE layer during decoding, isolating execution overhead (excluding routing) on Qwen3.}
    \label{fig:breakdown_comp}
\end{figure*}

\noindent\textbf{{Motivation 3: Sequential Execution in Resource-Constrained MoEs Amplifies Kernel Launch Overhead.}}
Limited GPU memory often forces MoE layers to execute experts nearly sequentially rather than in parallel, amplifying the inefficiencies of the naive expert-centric gather–compute–scatter process—referred to here as \textit{Bookkeeping}. 
By constructing a global routing mask and identifying token subsets for each expert, this procedure introduces highly scattered token access operations.
The problem is exacerbated in reasoning-optimized architectures. For example, Qwen3 employs top-8 routing (versus the typical top-2 in standard MoEs), which quadruples the number of kernel invocations devoted to scattered token-access operations. Consequently, a single forward pass through one MoE layer requires more than 128 kernel launches per token solely for expert execution, with scattered token access kernels accounting for 18.75\% of these calls.

Fig.~\ref{fig:breakdown_comp} compares \textit{Bookkeeping} and \textit{TokenRepacking} in a single MoE layer across batch sizes.
Under \textit{Bookkeeping}, kernel launch overhead, routing metadata management, and token access/layout transformation collectively account for 54.2\% of total execution time on average. With \textit{TokenRepacking}, this fraction drops to 40.0\%, driven by fewer token access/layout transformation kernels (–4.7\%) and lower kernel launch overhead (–6.7\%).
These inefficiencies motivate \textit{TokenRepacking}: grouping tokens by expert assignment enables batched expert execution, reduces scattered token access kernels, and improves efficiency in resource-constrained CoT reasoning.

\begin{figure}[t]
    \centering
    \includegraphics[width=1\linewidth]{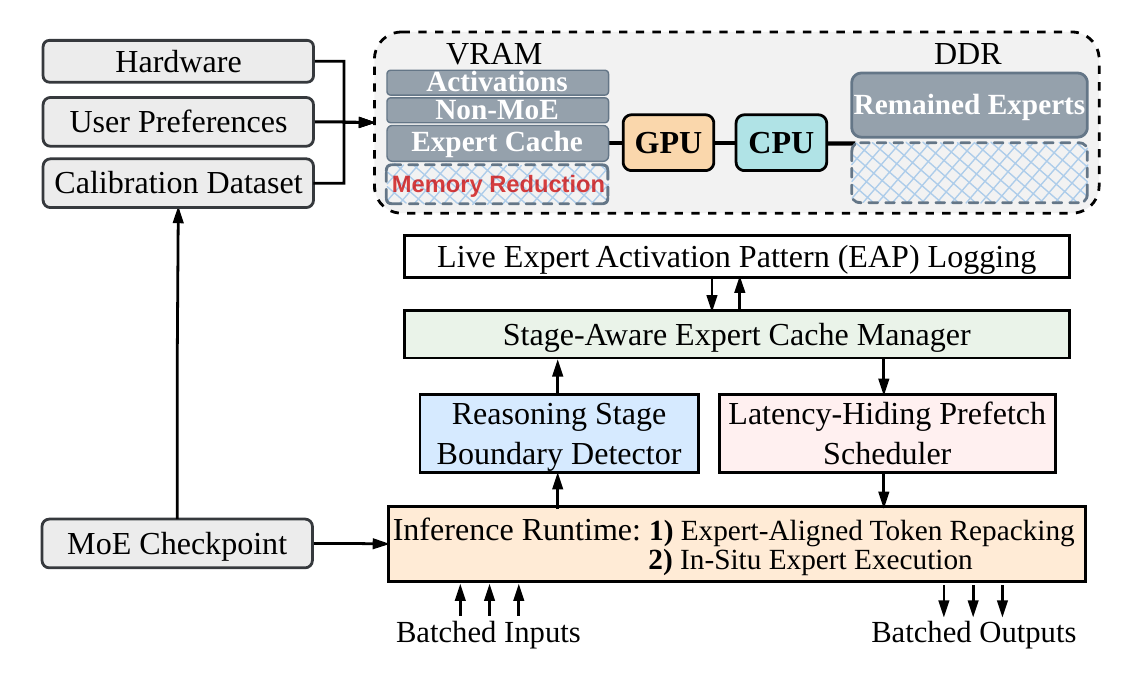}
    \caption{Design Overview of SAEM.}
    \label{fig:saem_design_flow}
\end{figure}

\section{SAEM: System Design}
\label{sec:system_design}
\subsection{Overview}
We now present SAEM, a data-aware MoE inference runtime that exploits stage-level regularities in CoT reasoning.
Fig.~\ref{fig:saem_design_flow} shows its overall architecture and data flow. SAEM comprises three core components: (i) a reasoning-stage boundary detector that identifies semantic transitions in CoT traces; (ii) a stage-aware expert cache manager that determines GPU--CPU expert placement based on historical activation patterns; and (iii) a latency-hiding prefetch scheduler that overlaps data movement with computation. In addition, SAEM incorporates two always-on optimizations: expert-aligned token repacking to improve memory coalescing and GPU utilization, and in-situ expert execution to avoid unnecessary weight transfers by executing CPU-resident experts directly on the host.

\subsection{Reasoning Stage Boundary Detection}
\subsubsection{Lightweight Pattern Matching for Stage Transitions} 
\label{subsubsec:pattern_match_stage_transitions}
\paragraph{Customized Transition Pattern Extraction} 
SAEM identifies stage boundaries from the generated token stream using lightweight lexical pattern matching. CoT traces are often organized into semantically coherent stages, such as initial derivation, alternative path exploration, self-correction, verification, and finalization. In this work, we focus on alternative path exploration, while extending detection to broader reasoning strategies is left for future work. These stage transitions are frequently indicated by discourse-level cues, such as ``Alternatively,'' ``Instead,'' ``Another way is,'' and ``On second thought,'' which provide a low-cost signal of shifts in the model's reasoning strategy.

SAEM avoids heavyweight semantic parsing or additional neural boundary classifiers, as they would introduce extra computation on the critical decoding path. Instead, cue-based detection only maintains a compact set of transition patterns and matches them against recently generated tokens. This design is sufficient for SAEM because the detector is not intended to produce perfect human-interpretable segmentation; rather, it only needs to identify coarse transition points where expert activation patterns are likely to change, enabling expert placement updates at stage granularity.

\begin{figure}[t]    
    \centering    
    \includegraphics[width=0.9\linewidth]{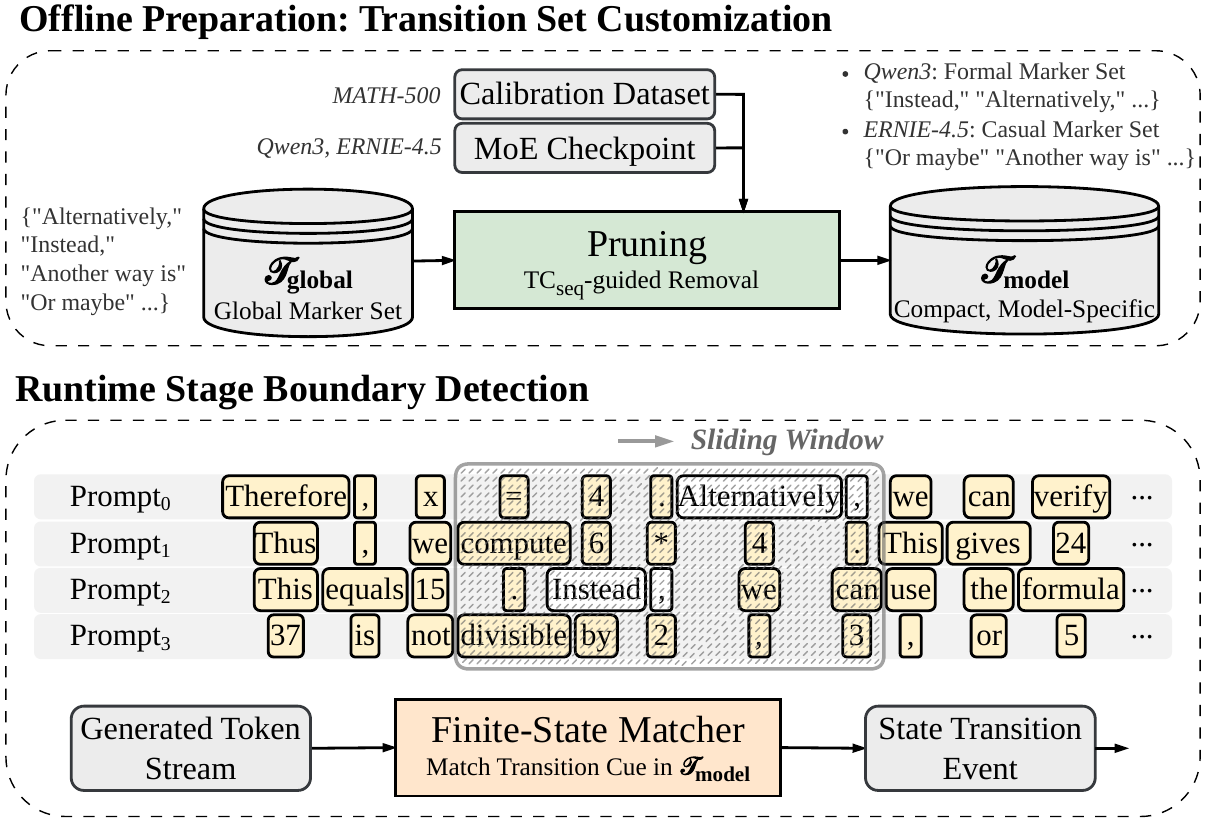}    
    \caption{Lightweight pattern matching for stage transitions.} 
    \label{fig:stage_boundary_detection}
\end{figure}

To build the transition patterns, SAEM first collects a global transition set $\mathcal{T}_{\text{global}}$ from common CoT traces. Directly using all markers in this set may increase matching overhead and introduce false positives. Moreover, different reasoning LLMs exhibit distinct stylistic tendencies shaped by their training and post-training pipelines. For example, Qwen3 tends to use more formal transition phrases, whereas ERNIE-4.5 often produces more conversational cues. 
Therefore, as illustrated in Fig.~\ref{fig:stage_boundary_detection}, SAEM customizes the transition set for each target MoE model. Given a calibration dataset and model checkpoint, SAEM evaluates candidate markers according to their contribution to stage-level expert activation coherence, measured by the temporal coherence metric $TC_{\text{seq}}$ introduced in Section~\ref{sec:motivation}. Markers that rarely appear, produce noisy segmentation, or do not improve adjacent-stage coherence are iteratively removed. This offline pruning yields a compact model-specific set $\mathcal{T}_{\text{model}}$, reducing runtime overhead while retaining high-impact cues associated with expert-activation shifts. 

\paragraph{Sliding-Window Pattern Matching}
At runtime, SAEM maintains a configurable sliding window over recently generated tokens for each active sequence. After each decoding step, the detector updates the window and checks whether its suffix matches any cue in $\mathcal{T}_{\text{model}}$. The matcher is implemented as a finite-state machine to track partial matches across token boundaries, which is necessary because transition cues may span multiple subword tokens. Once a complete cue is matched, SAEM emits a stage transition event, finalizes the routing statistics of the completed stage, and starts collecting a new activation profile for the next stage. Thus, the detector provides a lightweight runtime trigger that links semantic shifts in CoT reasoning to coarse-grained expert management.

\subsubsection{Activation Pattern Aggregation at Boundaries} 
When a stage transition event is triggered, SAEM aggregates expert activation statistics for the completed stage across all queries in the batch. Specifically, it computes per-layer, per-expert usage frequencies from the observed routing decisions, summarizing the stage's overall computational demand rather than reacting to individual token-level fluctuations.

SAEM then uses the aggregated activation pattern to guide expert placement for the upcoming stage. Based on the temporal coherence described in Section~\ref{sec:motivation}, frequently activated experts in the completed stage are likely to remain useful in the adjacent stage. SAEM therefore prioritizes these experts for GPU residency and deprioritizes rarely activated ones. By updating the cache only at detected stage boundaries, SAEM avoids the overhead of per-token expert migration while preserving adaptivity to structured changes in the reasoning trajectory.

\subsection{Stage-Aware Expert Cache Management}

\subsubsection{Memory Initialization and Allocation} 
At system startup, non-MoE layers, including attention mechanisms and layer normalization, are placed directly on the GPU due to their small footprint and consistent activation across all tokens. For MoE layers, SAEM allocates a fixed-size GPU memory region for expert weights, with the cache capacity determined by the user-specified Expert Cache Ratio (ECR). This fixed allocation ensures predictable memory usage throughout inference and prevents memory fragmentation that could arise from dynamic allocation. Optionally, the system can warm the cache with a default hot set of experts identified through dataset-level calibration on representative reasoning tasks. The ECR-defined cache budget is evenly distributed across layers and filled first with the highest-priority experts; any remaining capacity is assigned to globally frequent experts. Once finalized, the per-layer cache slot layout remains fixed.
This initialization provides a strong starting point, but expert placement is continuously refined at runtime as stage-specific activation patterns emerge during actual inference.

\subsubsection{Dynamic Expert Placement Using Historical Patterns} 
As inference progresses, SAEM maintains two key data structures: a live activation log that tracks the cumulative activation frequency of each expert within the current reasoning stage, and a residency map indicating which experts occupy the fixed per-layer GPU cache slots.
At each stage boundary, the residency planner updates expert placement within each layer’s allocated cache budget. Experts in a given layer are ranked by their activation frequency in the just-completed stage, and the least-used experts in that layer’s cache slots are evicted to make room for higher-priority experts currently residing on the CPU. Only replacements compatible with the fixed per-layer slot layout are permitted, preserving the memory organization defined at initialization. After the update, the activation log is reset to begin collecting statistics for the next stage.
This layer-local, stage-aware placement strategy leverages the observed temporal coherence: consecutive reasoning stages exhibit over 90\% overlap in activated expert subsets, allowing SAEM to maintain an efficient and predictive caching policy.

\subsubsection{Latency-Hiding Prefetch and Overlap Optimization} 
SAEM reduces expert weight transfer latency by overlapping data movement with computation. 
Expert weight transfers between CPU and GPU are orchestrated using asynchronous DMA operations that overlap with ongoing kernel execution.
Lightweight GPU event mechanisms coordinate these asynchronous operations by signaling transfer completion, ensuring that the required expert weights become available exactly when needed without stalling the pipeline.
Upon detecting a stage boundary, the scheduler immediately issues prefetch requests, allowing weight transfers to proceed in parallel with the final decoding steps of the current stage and the initial computations of the next stage. This prefetch-and-compute overlap minimizes cache update costs and helps maintain high GPU utilization throughout inference. 
Crucially, SAEM performs expert migration exclusively at stage boundaries rather than at token granularity, reducing scheduling overhead and aligning data movement with natural transitions in expert activation patterns.

\subsection{In-Situ CPU Expert Execution}
\subsubsection{Avoiding Unnecessary Data Movement via Local Execution} A key insight in SAEM is that not every expert must reside on the GPU. 
When a requested expert is not GPU-resident but remains available in host memory, SAEM compares two execution paths: transferring the expert to the GPU for execution or executing it directly on the CPU. It then selects the lower-latency option for that expert. CPU execution is often preferable when only a few tokens are assigned to the expert, as PCIe transfer latency dominates in this regime. As the expert-specific token count increases, however, GPU execution becomes more favorable because the transfer cost is amortized over more tokens. In this way, CPU execution avoids the true bottleneck---PCIe transfer latency---rather than introducing one. Moreover, migrating infrequently activated experts to the GPU would unnecessarily pollute the cache and risk evicting more valuable, frequently used experts.

\subsubsection{Hybrid CPU--GPU Execution Pipeline} The in-situ execution workflow proceeds as follows. After the routing mechanism assigns tokens to experts under the top-$k$ gating policy, SAEM partitions the token batch into GPU-bound and CPU-bound groups according to the current expert residency map. GPU-resident experts are executed using high-throughput GPU kernels that exploit tensor-core acceleration, while CPU-resident experts are processed in parallel using optimized GEMM backends such as Intel MKL or OpenBLAS. 

To reduce overheads associated with remote memory access and scheduling interference, SAEM supports NUMA-aware CPU execution by co-locating expert computation threads and their associated data within the same CPU socket. CPU-resident experts are executed by threads pinned to cores on a single socket, and expert weights and activation buffers are allocated on the corresponding NUMA node. This locality-aware execution model minimizes cross-socket memory traffic and reduces access latency during hybrid GPU--CPU inference.

The outputs from GPU and CPU execution paths are merged through a lightweight synchronization step before advancing to the next layer. This hybrid strategy offers two key benefits: it avoids unnecessary transfers for infrequently activated experts that would otherwise pollute the GPU cache, and it improves overall throughput by using CPU compute resources alongside the GPU—transforming the CPU from a passive storage device into an active computation engine.

\subsection{Expert-Aligned Token Repacking}

\subsubsection{Expert-Aligned Token Repacking} Motivated by the inefficiencies identified in Section~\ref{sec:motivation}, SAEM introduces a token repacking mechanism that reorganizes tokens by expert assignment in each MoE layer. 
By grouping tokens by expert IDs and placing them contiguously in memory, repacking removes the scattered token access operations inherent in the naive \textit{Bookkeeping} approach, which repeatedly gathers non-contiguous tokens and launches fragmented kernels. 
The temporary workspace required for repacking is allocated once and reused across all MoE layers, and the cost of grouping tokens by expert is negligible because it involves only lightweight reindexing over the current batch, resulting in minimal overhead even for deep models.

\begin{figure*}[t]
    \centerline{\includegraphics[width=0.8\linewidth]{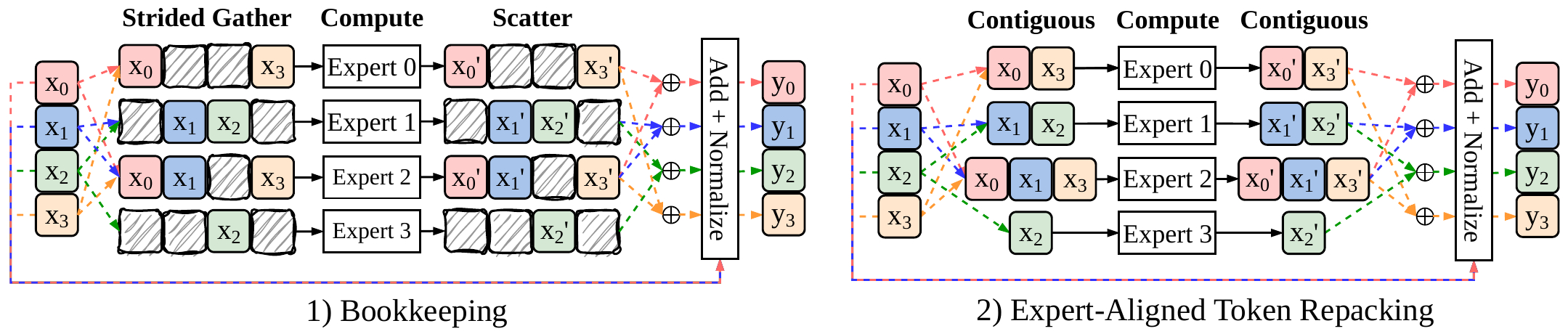}}
    \caption{Illustration of expert-aligned token repacking for a single MoE layer during decoding with a batch size of four. For clarity, the routing decision is assumed but not shown.}
    \label{fig:token_repacking}
\end{figure*}

This alignment enables each expert to execute its computation in a single batched kernel launch rather than multiple small invocations, directly reducing kernel launch overhead and layout transformation costs (Fig.~\ref{fig:breakdown_comp}). 
Fig.~\ref{fig:token_repacking} illustrates the mechanism. Consider four tokens with top-2 routing: $x_0$ is routed to experts $E_0$ and $E_2$, $x_1$ to $E_1$ and $E_2$, $x_2$ to $E_1$ and $E_3$, and $x_3$ to $E_0$ and $E_2$. 
Without packing, token–expert pairs are scattered across memory, requiring each expert to extract its tokens via separate layout transformation kernels followed by multiple fragmented launches.
Token repacking removes this fragmentation by placing all tokens assigned to a given expert contiguously—for instance, $x_0$ and $x_3$ for $E_0$—allowing the expert to operate on a single, dense input buffer. 

\subsubsection{Applicability and Performance Impact}
Token repacking is applied at every MoE layer and operates independently of CoT stage boundaries. By converting sparse token-to-expert mappings into dense expert-aligned batches, it simplifies memory access, reduces layout transformation overhead, and decreases kernel launch frequency—yielding higher GPU throughput, particularly at small batch sizes or imbalanced routing.

\setlength\tabcolsep{4.2pt} 
\begin{table*}[t]
    \centering
    \caption{Architectural details of MoE models optimized for multi-step reasoning tasks.}
    \label{tab:saem_moe_models}
        \begin{tabular}{|l|c|c|c|c|c|c|}
        \hline
            \multicolumn{1}{|c|}{Model} & \#Blocks & \#Shared Experts& \#Routed Experts & Top-k & \multicolumn{1}{c}{Experts Params.} & \multicolumn{1}{|c|}{Total Params.}\\
            \hline
            \hline
            Qwen3-30B-A3B (Qwen3) & 48 & 0 & 128 & 8 & 29.0B & 30.5B\\
            \hline
            ERNIE-4.5-21B-A3B (ERNIE-4.5) & 28 & 2 & 64 & 6 & 21.0B & 21.8B\\
        \hline
        \end{tabular}
\end{table*}

\begin{figure*}[t]
    \centering
    \begin{adjustbox}{center}
        \input{diagrams/speed_mem_comp_math}
    \end{adjustbox}
    \caption{Throughput comparison over batch sizes and ECRs for \textbf{MATH-500}. Overlaid lines denote speedup relative to the strongest baseline.}
    \label{fig:saem_throughput_comp_math}
\end{figure*} 

\begin{figure*}[t]
    \centering
    \begin{adjustbox}{center}
        \input{diagrams/speed_mem_comp_aime}
    \end{adjustbox}
    \caption{Throughput comparison over batch sizes and ECRs for \textbf{AIME 2024}. Overlaid lines denote speedup relative to the strongest baseline.}
    \label{fig:saem_throughput_comp_aime}
\end{figure*} 

\begin{figure*}[t]
    \centering
    \begin{adjustbox}{center}
        \input{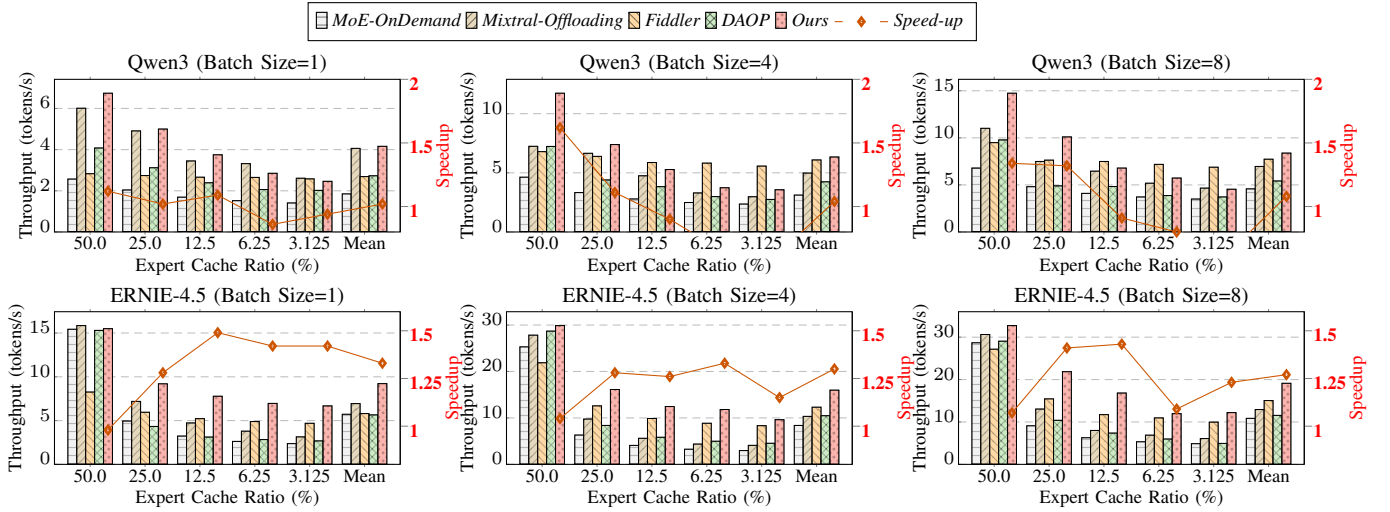}
    \end{adjustbox}
    \caption{Throughput comparison over batch sizes and ECRs for \textbf{GPQA-Diamond}. Overlaid lines denote speedup relative to the strongest baseline.}
    \label{fig:saem_throughput_comp_gpqa}
\end{figure*} 


\section{Experimental Evaluation}
\subsection{Experimental Setup}
\subsubsection{Models and Datasets} 
We evaluate SAEM on two reasoning-oriented MoE models: Qwen3-30B-A3B~\cite{yang2025qwen3} (Qwen3) and ERNIE-4.5-21B-A3B-Thinking~\cite{ernie2025technicalreport} (ERNIE-4.5). Their model configurations, including expert count and routing strategy, are summarized in Table~\ref{tab:saem_moe_models}.
We use MATH-500~\cite{hendrycks2021measuring} as the primary benchmark, which contains 500 Olympiad-style mathematics problems spanning seven domains. To assess generality beyond this setting, we also evaluate on AIME 2024~\cite{AIME2024}, a compact but challenging set of 30 competition-level mathematics problems covering topics such as algebra, geometry, number theory, combinatorics, and probability, and GPQA-Diamond~\cite{rein2024gpqa}, a 198-question graduate-level multiple-choice benchmark covering physics, chemistry, and biology. Together, these datasets evaluate SAEM across mathematical reasoning and scientific question answering, allowing us to assess both performance and temporal coherence of expert activations across diverse knowledge domains.

\subsubsection{Hardware} 
Experiments are conducted on a single-node setup equipped with a single NVIDIA A100 GPU (80~GB HBM2e, 1.93~TB/s). We use an Intel Xeon Gold 6326 CPU (16 cores, 2.9~GHz) to emulate practical deployment scenarios outside of data centers. The host system provides 512~GB DDR4 memory, with CPU--GPU communication over PCIe~4.0$\times$16 (64~GB/s). To ensure stable and reproducible performance measurements, CPU-based expert execution is confined to a single CPU socket, with threads pinned to cores within that socket and memory allocations restricted to the corresponding NUMA node. 
This configuration avoids cross-socket memory accesses and minimizes variability from OS scheduling and remote NUMA traffic, enabling isolation of SAEM’s algorithmic behavior from hardware-induced noise.

\subsubsection{Implementation}
We implement SAEM atop the Hugging Face Transformers library~\cite{wolf-etal-2020-transformers} using its PyTorch backend. We vary the batch size from 1 to 8. To evaluate the trade-off between memory and performance, we adopt the \textbf{Expert Cache Ratio (ECR)} metric~\cite{zhang2025daop}, defined as the ratio of GPU-resident experts to the total number of routed experts. Expert caches are initialized using statistics of dominant experts gathered from MATH-500. Inference efficiency is measured by end-to-end throughput, reported in tokens per second.

\subsubsection{Baselines}
We compare SAEM with several representative MoE inference baselines: MoE-OnDemand, Mixtral-Offloading~\cite{eliseev2023fast}, Fiddler~\cite{kamahori2024fiddler}, and DAOP~\cite{zhang2025daop}.
MoE-OnDemand keeps non-MoE layers (e.g., attention and normalization) and dominant experts on the GPU, while storing others in CPU memory. Non-resident experts are fetched on demand, often incurring substantial data transfer overhead.
Mixtral-Offloading employs an LRU-style token-level cache, dynamically migrating frequently used experts between CPU and GPU.
Fiddler reduces data movement by executing non-resident experts directly on the CPU whenever they are activated.
DAOP partitions experts between CPU and GPU based on per-sequence activation patterns and predicts future expert usage to precompute selected experts on the CPU.

\subsection{Speedup}
\label{sec:speedup}
Fig.~\ref{fig:saem_throughput_comp_math}, Fig.~\ref{fig:saem_throughput_comp_aime}, and Fig.~\ref{fig:saem_throughput_comp_gpqa} report end-to-end throughput of SAEM against state-of-the-art baselines across varying batch sizes and ECRs on MATH-500, AIME 2024, and GPQA-Diamond, respectively. 
SAEM outperforms the strongest baseline in nearly every configuration tested, with average speedups of \textbf{1.60$\times$} (Qwen3) and \textbf{1.47$\times$} (ERNIE-4.5) on MATH-500, \textbf{1.20$\times$} and \textbf{1.21$\times$} on AIME 2024, and \textbf{1.14$\times$} and \textbf{1.34$\times$} on GPQA-Diamond.

The margin is largest for MATH-500, the dataset used to calibrate both components of SAEM's offline preparation: the dominant-expert statistics that warm the initial cache and the pruned transition set $\mathcal{T}_{\text{model}}$. AIME 2024 and GPQA-Diamond are therefore held out, and their results characterize SAEM under calibration mismatch rather than under matched conditions. GPQA-Diamond additionally shifts domain, from mathematical derivation to graduate-level multiple-choice science, so its discourse-cue distribution differs from the distribution used to prune $\mathcal{T}_{\text{model}}$. That SAEM retains a 1.14--1.34$\times$ advantage with no target-domain calibration indicates the stage-locality mechanism transfers across reasoning domains, while the MATH-500 margin indicates the additional headroom available when calibration data is representative of the workload.
In the remainder of this section, we examine MATH-500 in detail, as it spans the widest range of cache-pressure conditions and isolates SAEM's behavior when calibration is representative; unless otherwise noted, the following analysis refers to Fig.~\ref{fig:saem_throughput_comp_math}.

\subsubsection{Single-batch Regime}
In the single-batch setting, aggressive LRU-driven token-level expert migration allows Mixtral-Offloading to maintain high GPU utilization by frequently refreshing the active expert set.
In contrast, SAEM updates its cache only at CoT boundaries, resulting in a slightly lower steady-state cache hit rate (e.g., 90.08\% vs.\ 94.53\% under 50\% ECR on Qwen3).
However, SAEM offsets this disadvantage with expert-aligned token repacking, which recovers tensor-core efficiency lost to irregular expert placement. As a result, SAEM still achieves a \textbf{1.62$\times$} throughput improvement over Mixtral-Offloading despite less frequent cache updates.

\subsubsection{Concurrent Multi-batch Regime}
Under multi-batch execution, the performance gap widens further in favor of SAEM. Fine-grained LRU migration incurs substantial redundant weight transfers under multi-batch execution, an overhead that grows rapidly as batch size increases and cache capacity becomes constrained.
SAEM avoids these inefficiencies through stage-aware expert caching, which selectively retains experts aligned with the CoT reasoning structure. This strategy consistently yields higher cache hit ratios and improved GPU utilization. 
For instance, on Qwen3 with batch size 8, SAEM increases the cache hit ratio relative to Mixtral-Offloading by 53.06\% at 12.5\% ECR and by \textbf{187.95\%} at 3.125\% ECR.
These improvements translate into tangible runtime benefits: SAEM achieves a \textbf{2.10$\times$} throughput gain over Mixtral-Offloading at batch size 8 and 3.125\% ECR.
Importantly, SAEM sustains its advantage even under low-ECR conditions, where the active expert set is small and cache thrashing severely limits LRU-based methods.

\subsubsection{Comparison with DAOP and Fiddler}
These improvements enable SAEM to outperform DAOP and Fiddler by 2.70$\times$ and 1.60$\times$ on average across all tested models and batch sizes on MATH-500. 
DAOP uses prediction-based expert precomputation to accelerate CPU-side expert execution and reduce CPU--GPU synchronization overhead, while Fiddler executes non-resident experts on the CPU whenever they are needed. Both approaches construct GPU caches using sentence-level statistics or calibration-dataset statistics to identify dominant experts.
However, neither method leverages semantic continuity across CoT reasoning stages. Their expert-selection mechanisms are static at the sentence level or calibration-dataset level and thus cannot adapt dynamically as reasoning unfolds. Consequently, both DAOP and Fiddler often fail to retain experts that are likely to be reused across multi-step reasoning, leading to weak cache reuse and significantly lower throughput compared with SAEM.

\subsection{Prediction-Oracle Upper-Bound Analysis}
\label{subsec:saem_prediction_oracle}

SAEM performs prediction-guided cache updates: at each detected stage boundary, it uses the expert-activation pattern (EAP) of the completed stage as a proxy for the upcoming stage. To isolate the performance headroom associated with this one-stage prediction lag, we construct a clairvoyant oracle that replaces the predictive proxy with perfect knowledge of the upcoming stage while preserving the cache capacity, update events, and cache-enforcement mechanism.

Let $\mathrm{EAP}_{s}$ denote the per-layer expert activation counts observed during stage $s$. We first execute the model normally and record the generated token sequence, stage-boundary events, and $\mathrm{EAP}_{s}$ for every stage. We then replay the recorded sequence under two policies. The online predictive policy updates the cache for stage $s$ using $\mathrm{EAP}_{s-1}$, whereas the oracle uses the recorded $\mathrm{EAP}_{s}$. Both policies enforce the same ECR and invoke the same cache-update and expert-execution mechanisms at identical boundary events; they differ only in the activation profile supplied to the residency planner. Because expert routing is logically independent of physical cache placement, the recorded trace can be reused across the evaluated ECRs.

Fixed-sequence replay is necessary for a controlled comparison. Atomic accumulation in the MoE combine operation can introduce small run-to-run numerical differences, causing unconstrained autoregressive generation to diverge in its output tokens and, consequently, its routing and stage boundaries. Replaying the recorded token sequence fixes the prompts, decoding length, and boundary events across the online predictive and oracle executions. We additionally verify that replayed cue events match the recorded events and disable profiling instrumentation during timing. This protocol isolates the effect of prediction quality while holding cache capacity and update cadence fixed.

We quantify the distance to the oracle using cache-hit-ratio efficiency and throughput efficiency:
\begin{equation}\eta_{\mathrm{CHR}}=\frac{\mathrm{CHR}_{\mathrm{SAEM}}}{\mathrm{CHR}_{\mathrm{Oracle}}},\qquad\eta_{\mathrm{TP}}=\frac{T_{\mathrm{SAEM}}}{T_{\mathrm{Oracle}}},\label{equ:upper_bound}\end{equation}
where $\mathrm{CHR}$ denotes the expert cache hit ratio and $T$ denotes end-to-end decoding throughput. Values approaching $100\%$ indicate that the online predictive policy performs close to the prediction oracle. Both terms in each ratio are measured using the same fixed-sequence replay protocol. We report ratios rather than the individual replay throughputs because the latter are not directly comparable to the free-running results in Fig.~\ref{fig:saem_throughput_comp_math}. For context, Table~\ref{tab:saem_prediction_oracle} also reports the ECR~=~100\% fully GPU-resident reference and SAEM's free-running throughput.

\setlength\tabcolsep{4.2pt} 
\begin{table}[t]
    \centering
    \caption{Prediction-oracle analysis on Qwen3. Full GPU is the ECR~=~100\% reference, and SAEM reports the free-running throughput from Fig.~\ref{fig:saem_throughput_comp_math}. CHR and TP Eff. are defined in Eq.~\eqref{equ:upper_bound}.}
    \label{tab:saem_prediction_oracle}
        \begin{tabular}{|c|c|c|r|r|r|}
            \hline
            \makecell{Batch\\Size} &
            \makecell{Full GPU\\(tokens/s)} &
            \makecell{ECR} &
            \makecell{SAEM\\(tokens/s)} &
            \makecell{CHR Eff. (\%)} &
            \makecell{TP Eff. (\%)} \\
            \hline
            \hline
            
            \multirow{3}{*}{1} &
            \multirow{3}{*}{9.04} &
            50.0\% & 7.19 & {90.36} & {87.87} \\
            \cline{3-6}
            & & 25.0\% & 6.45 & {81.40} & {90.82} \\
            \cline{3-6}
            & & 12.5\% & 6.10 & {76.09} & {92.80} \\
    
            \hline
    
            \multirow{3}{*}{8} &
            \multirow{3}{*}{20.90} &
            50.0\% & 18.55 & {97.80} & {95.69} \\
            \cline{3-6}
            & & 25.0\% & 16.35 & {96.43} & {98.46} \\
            \cline{3-6}
            & & 12.5\% & 13.82 & {96.75} & {100.78} \\
    
            \hline
        \end{tabular}
\end{table}

Table~\ref{tab:saem_prediction_oracle} measures SAEM's distance from the prediction oracle in cache placement and end-to-end performance. At batch size 1, CHR efficiency decreases from 90.36\% to 76.09\% as ECR decreases, indicating that perfect prediction becomes increasingly beneficial for cache placement under tighter memory constraints. In contrast, TP efficiency increases from 87.87\% to 92.80\%, limiting the oracle throughput improvement to $1.08$--$1.14\times$. This divergence shows that higher cache accuracy does not translate proportionally into throughput because only part of the end-to-end latency is affected by expert residency.

At batch size 8, CHR and TP efficiencies remain above 96\% and 95\%, respectively, indicating that SAEM already operates close to the oracle under higher concurrency. The 100.78\% TP efficiency at ECR~=~12.5\% reflects sub-1\% measurement variation and is treated as parity. Overall, perfect prediction provides the greatest cache-placement improvement at low ECR and batch size 1, but only modest throughput gains. The substantially larger gap between SAEM and Full GPU therefore arises primarily from limited GPU cache capacity rather than prediction inaccuracy.

\setlength\tabcolsep{4.2pt} 
\begin{table}[t]
    \centering
    \caption{Inference speedup breakdown of proposed techniques under 12.5\% cache ratio on Qwen3.}
    \label{tab:speedup_breakdown}
        \begin{tabular}{|c|l|r|r|}
        \hline
        \makecell{Batch\\Size} & \multicolumn{1}{c}{Technique} & \multicolumn{1}{|c|}{\makecell{Throughput\\(tokens/s)}} & \multicolumn{1}{c|}{Speedup} \\
        \hline
        \hline
        \multirow{4}{*}{1} 
            & Best-performing baseline        & 3.45 & -- \\
            \cline{2-4}
            & Cache update only   & 4.81 & 1.39$\times$ \\
            \cline{2-4}
            & In-situ CPU execution only & 2.73  & 0.79$\times$ \\
            \cline{2-4}
            & Token repacker only             & 5.02  & 1.46$\times$ \\
            \cline{2-4}
            & \cellcolor{orange!30}All                              & \cellcolor{orange!30}6.10 & \cellcolor{orange!30}1.77$\times$ \\
        \hline
        \multirow{4}{*}{8}
            & Best-performing baseline        & 8.51  & -- \\
            \cline{2-4}
            & Cache update only   & 10.44 & 1.23$\times$ \\
            \cline{2-4}
            & In-situ CPU execution only & 8.51  & 1.00$\times$ \\
            \cline{2-4}
            & Token repacker only             & 9.09 & 1.07$\times$ \\
            \cline{2-4}
            & \cellcolor{orange!30}All                              & \cellcolor{orange!30}13.82 & \cellcolor{orange!30}1.62$\times$ \\
        \hline
        \end{tabular}
\end{table} 

\subsection{Ablation Study}
We perform an ablation study to quantify the contribution of each SAEM component. Inference speedups are measured on Qwen3 at 12.5\% ECR and reported relative to the strongest baseline, as summarized in Table~\ref{tab:speedup_breakdown}. Stage-aware cache updates and expert-aligned token repacking each accelerate inference in isolation, addressing complementary bottlenecks: the former retains experts likely to be reused across adjacent reasoning stages, reducing cache churn and expert migration, while the latter groups routed tokens into contiguous, hardware-friendly batches, improving memory locality and kernel efficiency. In-situ CPU execution alone, however, reaches only 0.79$\times$ at batch size 1 and merely matches the baseline at batch size 8: without stage-aware placement, the GPU cache retains a suboptimal expert set, forcing many tokens onto the slower CPU path---effectively Fiddler-style execution without the mechanisms that make hybrid execution profitable. It is thus an enabling mechanism rather than a standalone accelerator. Combining all three components yields the highest throughput (1.77$\times$ and 1.62$\times$), confirming that the mechanisms are complementary.

To isolate the contribution of in-situ execution \emph{within} the full system, we disable CPU-side expert execution while preserving the stage-aware cache policy and token repacking; non-resident experts are instead transferred to the GPU on demand through a reserved temporary slot. Throughput drops from 6.10 to 3.22 tokens/s at batch size 1 and from 13.82 to 9.00 tokens/s at batch size 8, confirming that avoiding on-demand PCIe transfers remains essential under constrained GPU memory. Together with the isolation results, this shows that SAEM's gains arise from coordinating hybrid CPU--GPU execution with stage-aware expert placement and kernel-efficient token organization, rather than from any single mechanism.

\section{Discussion \& Limitations}
SAEM improves MoE inference throughput by exploiting reasoning-stage locality for expert management. We discuss its runtime overhead and remaining limitations below.

\subsection{Practical Overhead}
SAEM introduces runtime overhead from transition-cue matching, token repacking, and stage-level expert placement. These costs are bounded by design and remain lightweight relative to MoE computation and data movement. Transition detection performs pattern matching over a short window of recently generated tokens, avoiding additional neural classifiers or semantic parsing during decoding. Token repacking uses index-based reordering with preallocated workspaces, avoiding repeated memory allocation and limiting layout-transformation overhead. Expert placement is updated only at detected stage boundaries rather than at every decoding step, amortizing cache management cost over the following reasoning stage. As shown in the ablation study, these costs are outweighed by reductions in PCIe expert transfers, cache churn, and fragmented token-access kernels, resulting in net throughput improvement.

\subsection{Stage Boundary Detection}
SAEM currently detects reasoning-stage boundaries using explicit linguistic transition cues in generated CoT traces, such as ``Alternatively,'' and ``Instead,''. This strategy is effective when reasoning traces contain clear structural markers. However, some queries may lack explicit cues due to variation in generation style or task-specific reasoning patterns. In such cases, boundary detection may become less precise, reducing the effectiveness of stage-level cache updates.

To improve robustness beyond explicit textual cues, future work could explore statistical boundary signals derived from expert activations. One promising direction is to track changes in the entropy of expert activation distributions across successive tokens or short windows. Pronounced entropy shifts may indicate implicit transitions between reasoning stages even when discourse markers are absent. Such signals could complement lexical cues and improve SAEM's generality across models, reasoning styles, and workloads.

\section{Conclusion}
SAEM demonstrates that semantics-aware execution strategies for reasoning workloads can significantly improve the efficiency of MoE inference for CoT reasoning. By leveraging stage-level activation patterns and applying coordinated caching, scheduling, and token repacking, SAEM reduces memory pressure and accelerates decoding. These results highlight the potential of structurally informed runtime designs in advancing scalable and efficient CoT reasoning under resource constraints.

\bibliographystyle{IEEEtran}
\bibliography{refs}

\end{document}